\documentclass{article}
\usepackage{spconf,amsmath,graphicx,booktabs}
\usepackage{microtype,xurl}
\usepackage{placeins}
\usepackage{float}
\usepackage{hyperref}
\title{Monsoon: Scalable Spontaneous Speech Collection for Low-Resource ASR in 50 Languages}
\name{\parbox{0.9\textwidth}{\centering Shobhit~Banga, Manas~Dhir, Aaditya~Pareek, Amritansh~Walecha, Bhaskar~Singh, Hanuman~Sidh, Manmeet~Kaur, Sagar~Jain, Supriya~Paul, Vanshika~Chhabra, Utkarsh~Singh}}
\address{Voice Arena \\ \texttt{\{shobhit, manas, aaditya\}@voicearena.com}}

\begin{document}
%
\maketitle
\begin{abstract}
For most languages, automatic speech recognition (ASR) is limited less by
model architecture than by a lack of accurately transcribed spontaneous
speech. We present Monsoon, a framework for collecting such data at scale. It
elicits unscripted conversations between paired contributors, screens
recordings with automated classifiers, corrects transcripts over five levels
of human review, and selects segments for lexical and acoustic coverage. It has
collected speech in 50 languages; we analyse seven 3,500-hour corpora. At equal volume, its quality stages
reduce semantic word error rate (WER) by 10--39\% relative to a conventional protocol.
Compared with an architecture-matched baseline trained on public corpora,
Whisper fine-tuned on Monsoon performs comparably on the baseline's own test
sets and reduces semantic WER by 20--44 points on conversational benchmarks. The gains
transfer to a second architecture and grow with data volume.
\end{abstract}

\section{Introduction}
\label{sec:intro}

For English, where transcribed speech is abundant, the best ASR systems now average below 6\% word error rate (WER) across
the Open ASR Leaderboard~\cite{srivastav2025openasr}. Across languages,
accuracy tracks the quantity of such data; for Whisper, WER decreases
log-linearly with the hours of training audio per language, halving with every
sixteen-fold increase~\cite{radford2022whisper}. For most languages, accuracy
is therefore largely limited by the amount of suitable training data rather
than by model architecture. Existing corpora for these languages are small,
dominated by read or prompted speech that is mismatched to spontaneous usage,
or weakly verified: annotation guidelines vary within and between corpora,
crowd-sourced transcripts are rarely checked, and inconsistent orthography
fragments the label space~\cite{taguchi2024language}. Much of this noise is
unsystematic, so text normalisation cannot fully correct it.

Efforts to scale speech data have mostly favoured volume over control.
Crowd-sourced read speech~\cite{ardila2020}, parliamentary and
religious recordings~\cite{wang2021,pratap2023}, and weakly supervised web
audio~\cite{radford2022whisper} give broad language coverage, but the source
fixes their domain, speaking style, and label quality. Models trained on large
aggregations of such data do well on test sets drawn from the same collections
and lose much of their accuracy on conversational
speech~\cite{bhogale2023,bhogale2026voiceindialargescalebenchmark}. What is
scarce, then, is not speech data in general but data collected and verified for
the conditions a deployed model faces.

We propose \emph{Monsoon}, a framework that enforces data quality at scale
through automated filtering and human verification. The framework has been
used to collect speech in 50 languages. Here we analyse seven of them,
including Kashmiri, which has no public training corpus of comparable size,
each with about 3{,}500 hours of transcribed spontaneous speech in
single-speaker segments taken from the separately recorded channels of
two-party conversations. Per language, these corpora exceed publicly
available corpora~\cite{javed2024indicvoices,indicvoices_hf} in transcribed
volume and, on average, in speaker coverage. Whisper
medium~\cite{radford2022whisper} fine-tuned
on Monsoon matches the architecture-matched IndicWhisper~\cite{bhogale2023} on
the latter's own test sets and outperforms it by 20 to 44 points of semantic
WER on independently collected spontaneous speech. At equal volume
and with the same elicitation, the framework's quality stages reduce error by
10 to 39\% relative over a conventional protocol. The gains transfer to
SraVaani-1.0~\cite{sravaani2026}, an independently pretrained, Indic-specialised
FastConformer model; the fine-tuned Whisper model
outperforms all but one of five commercial systems; and error is still falling
at the full collected volume. A 10-hour sample of each of the 50 languages is
available on request.

\section{Related Work}
\label{sec:related}

\noindent\textbf{Conversational speech corpora.} Transcribed spontaneous
speech has long been expensive to collect. Switchboard~\cite{godfrey1992switchboard} established telephone
conversation as a training resource for English, and the Babel
programme~\cite{gales2014babel} extended it to low-resource languages at tens
of hours per language. IndicVoices~\cite{javed2024indicvoices,indicvoices_hf}
collects extempore and conversational speech across 22 languages, with 11.2K
transcribed hours in total. Benchmarks of accented and in-the-wild speech
continue to find that existing systems perform poorly outside read
speech~\cite{javed2024lahaja,bhogale2026voiceindialargescalebenchmark}.

\noindent\textbf{Scaling speech data.} Large multilingual corpora draw on
sources that already exist: volunteer read speech~\cite{ardila2020},
parliamentary recordings~\cite{wang2021}, religious texts~\cite{pratap2023},
weakly supervised web audio~\cite{radford2022whisper}, or aggregations of
earlier public corpora~\cite{bhogale2023}. The source fixes their speaking
style and domain, and label quality is controlled, if at all, by filtering
after the fact.

\noindent\textbf{Data quality in ASR training.} Orthographic complexity
predicts ASR error across languages~\cite{taguchi2024language}, speaker count
at fixed volume drives robustness to unseen accents~\cite{yong2025speakercount},
and non-intrusive predictors such as DNSMOS~\cite{reddy2021dnsmos} measure
acoustic quality at corpus scale. These factors have been studied in isolation,
not combined in a collection process and tested against conventional
collection at equal volume.

\section{The Monsoon Framework}
\label{sec:monsoon}

\noindent\textbf{Contributors.} Monsoon recruits speakers per language from an
existing contributor network, including rural and semi-urban regions, and
admits only those who pass a proficiency screen in the target language.
Contributors are paid, give informed consent, and grant rights for model
training and distribution. A per-language cap on recorded speech keeps any
speaker below 0.2\% of a corpus.

\noindent\textbf{Elicitation.} The platform pairs two contributors, typically
strangers, and records their conversation peer-to-peer on their own phones,
one speaker per channel. Each session follows a topic: a one-sentence scenario
with five sub-topics revealed in turn, sustaining 10 to 20 minutes of unscripted
speech. Topics are generated by Gemini 3.7 Flash across 35 everyday domains,
conditioned on the topics already accepted in each. Sub-topics follow one of
six conversational arcs, and constraints require every stage to involve both
speakers, at least one stage to elicit numbers and one a procedure, and at
least two to require naming real places, shops, or products, while the prompt
itself stays region-neutral and excludes role-play. Candidates are embedded
with BGE-small and rejected if their cosine similarity to
the nearest accepted topic exceeds 0.8, a threshold tuned by inspecting
near-duplicate pairs; the rejection rate rises from 16\% to
over 50\% as a domain saturates. Accepted topics are translated and revised by
native speakers. The pool holds about 5{,}000 topics per language and grows
with recording, so no pair receives the same topic twice.

\noindent\textbf{Gating and segmentation.} Classifiers fine-tuned from XLS-R
300M~\cite{babu2022xlsr} on over 20{,}000 hours of labelled audio verify
language and speaker gender with over 95\% held-out accuracy, and
PANNs~\cite{kong2020panns} rejects recordings in which any recorded-media
class, such as music, exceeds probability 0.5. Each channel is segmented
separately with WebRTC VAD at pauses of at least 2\,s, with a
soft limit of 15\,s, so every segment has one speaker. Segments with DNSMOS
P.808~\cite{reddy2021dnsmos} below 2.6, between ``poor'' and ``fair'' on the
absolute category rating scale, are kept only if human reviewers judge them
intelligible, and a per-language Whisper medium
model~\cite{radford2022whisper}, fine-tuned on 1{,}000 hours of annotated
data, produces draft transcripts.

\noindent\textbf{Transcription.} Drafts pass five levels, each an edit by one
annotator verified by another: the first corrects the draft against the
audio, and later levels target ambiguous pronunciations, code-switch
boundaries, named entities, and spelling variants. Numbers are written as words, and segments still disputed after the
fifth level are re-transcribed. Annotators are audited automatically on
batches of 100 segments and flagged for manual review if a batch contains
out-of-script characters, repeats the same insertion across many segments, or
systematically raises the minimum CER against several open-source ASR outputs;
the thresholds are adjusted per language as annotator behaviour is observed. After all
validation steps, an in-house team audits a sample of each annotator's
segments, and that annotator's segments enter the corpus only if at least 98\% of the
sampled segments are judged correct.

\noindent\textbf{Selection.} Unfiltered conversations contain many
greetings, fillers, and backchannels, and their quality and length are uneven.
About 67\% of recorded audio survives gating and transcription, and each
corpus is selected from a verified pool roughly ten times its final size. Segments are binned by DNSMOS, with edges at 2.6,
3.0, and 3.4, and by duration, and ranked within each bin by the mean
inverse pool frequency of their words $w_1,\dots,w_n$,
\begin{equation}
  s(x) = \tfrac{1}{n}\textstyle\sum_{i=1}^{n} 1/f(w_i),
  \label{eq:score}
\end{equation}
which favours rare vocabulary independently of length and demotes
filler-dominated segments. Every bin is given the same target of hours and
filled in rank order, per speaker and gender, subject to the speaker cap;
bins with too little data contribute all they have, and the shortfall is
taken up by the remaining bins. Scarce conditions such as noisy and short audio
are therefore kept as a small but present tail rather than crowded out.

\begin{table}[t]
\centering
\caption{Corpus statistics. Dur.: median segment length (s); F: female
share of audio; Dist.: districts; Dev.: device models; MOS: median DNSMOS.}
\label{tab:corpus}
\small
\setlength{\tabcolsep}{2.6pt}
\begin{tabular}{@{}lrrrrrrrr@{}}
\toprule
Lang. & Hours & Utt. & Dur. & Spk. & F & Dist. & Dev. & MOS \\
\midrule
Bengali  & 3{,}500 & 1.04M & 13.4 & 4{,}863  & 62\% & 127 & 787     & 3.66 \\
Gujarati & 3{,}544 & 1.22M & 11.9 & 5{,}568  & 40\% & 106 & 806     & 3.53 \\
Hindi    & 3{,}611 & 1.16M & 12.9 & 13{,}731 & 37\% & 781 & 1{,}123 & 3.61 \\
Kashmiri & 3{,}330 & 1.62M & 6.4 & 2{,}936  & 19\% & 31  & 447     & 3.61 \\
Marathi  & 3{,}500 & 1.08M & 13.2 & 5{,}329  & 57\% & 111 & 812     & 3.63 \\
Tamil    & 3{,}488 & 1.14M & 12.5 & 3{,}452  & 58\% & 95  & 777     & 3.56 \\
Telugu   & 3{,}500 & 1.18M & 12.3 & 6{,}068  & 53\% & 134 & 834     & 3.54 \\
\midrule
Total    & 24{,}472 & 8.43M & -- & 41{,}947 & -- & -- & -- & -- \\
\bottomrule
\end{tabular}
\end{table}

\begin{figure}[t]
\centering
\includegraphics[width=\columnwidth]{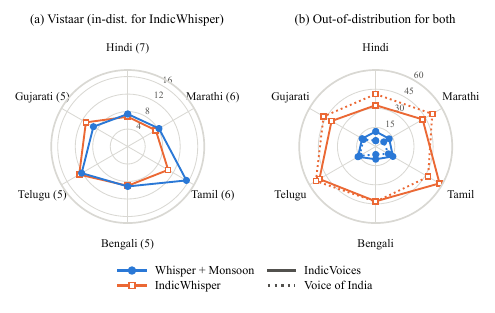}
\caption{Semantic WER (\%) of Monsoon-trained Whisper and IndicWhisper.
(a) Vistaar mean (sets per language in brackets); (b) IndicVoices (solid) and
Voice of India (dotted, OIWER). Radial scales differ.}
\label{fig:radar}
\end{figure}

\noindent\textbf{Resulting corpora.} Table~\ref{tab:corpus} summarises the
seven corpora. Hindi spans 781 districts, most of India; Kashmiri, drawn from
a far smaller population, has fewer and deeper contributors, shorter segments,
and the most skewed gender balance.

\noindent\textbf{Experimental setup.} We fine-tune Whisper
medium~\cite{radford2022whisper} separately on each Monsoon corpus, and
continually fine-tune SraVaani~\cite{sravaani2026}, with the same schedule:
five epochs of AdamW at a learning rate of $10^{-5}$, 500 warm-up steps, and
cosine decay, on eight H100 GPUs in bf16 with gradient clipping at 1.0, a per-device batch
of 16 with two accumulation steps (global 256), and SpecAugment. Whisper uses
time and frequency masking (probability 0.05, width 10); SraVaani uses weight
decay $10^{-3}$ and its default SpecAugment. We evaluate on the Vistaar test
sets~\cite{bhogale2023}, CommonVoice~\cite{ardila2020},
FLEURS~\cite{conneau2022}, GramVaani, IndicTTS, Kathbath and its noisy
variant~\cite{javed2023indicsuperb}, and MUCS~\cite{diwan2021}, which are in
distribution for the IndicWhisper baseline but not for Monsoon; and on
IndicVoices~\cite{javed2024indicvoices} and Voice of
India~\cite{bhogale2026voiceindialargescalebenchmark}, which are out of
distribution for both. We report semantic WER, in which an LLM judge
(Gemini 3.6 Flash) discounts differences that preserve the spoken
content~\cite{sarvam_llm_wer};
Voice of India is scored with its orthography-aware
variant~\cite{bhogale2026oiwer}. Kashmiri is excluded from the comparisons,
since none of the baseline or commercial systems supports it.

\section{Results}
\label{sec:results}

\subsection{Does Monsoon generalise beyond its own collection?}
\label{sec:iw}

We compare against IndicWhisper~\cite{bhogale2023}, a Whisper medium model
trained on about 10{,}700 hours of public corpora drawn from the same sources
as the Vistaar test sets. With the architecture held fixed, the comparison
mainly reflects the training data, although the two training recipes differ. Fig.~\ref{fig:radar} reports both
models per language.

On the Vistaar sets, which favour IndicWhisper because its training data
shares their sources, the two polygons largely coincide
(Fig.~\ref{fig:radar}a). The Monsoon model is ahead in Gujarati and Telugu and
within one point in Hindi, Marathi, and Bengali; only in Tamil does
IndicWhisper lead by more. Monsoon therefore matches a model trained on the
test sets' own sources without having seen any of them.

On the benchmarks out of distribution for both models, the polygons separate
completely
(Fig.~\ref{fig:radar}b). On both IndicVoices and Voice of India, every
IndicWhisper point lies above 32\% and every Monsoon point below 17\%, and
within each language the gap is at least 20 points. IndicWhisper's error grows
several-fold once the test data leaves its training sources, while Monsoon's
stays close to its level in panel (a). The gap reflects the domain of the
training data: IndicWhisper's largely read-speech sources do not prepare it for
spontaneous speech, which Monsoon is collected to capture, and the Monsoon model
transfers to independently collected spontaneous speech with little loss.

\subsection{What does the pipeline itself contribute?}
\label{sec:pipeline}

\begin{figure}[t]
\centering
\includegraphics[width=\columnwidth]{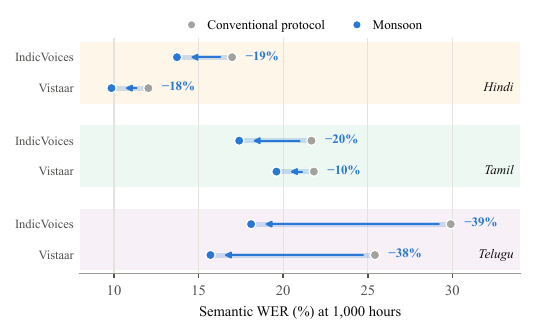}
\caption{Semantic WER (\%) at 1{,}000 hours: conventional protocol (grey)
vs.\ Monsoon (blue), same recipe. Labels: relative reduction.}
\label{fig:ablation}
\end{figure}

To isolate the pipeline, we compare against data from the same elicitation
process, with the same topics, pairing, and devices, but processed under a
conventional protocol: recordings are checked only for spoken language and
receive a single level of human transcription instead of five, with no further
gating, quality review, or selection. Both are Whisper medium
trained on 1{,}000 hours with the same recipe, so they differ only in how the
data was gated, transcribed, and selected. The
Monsoon model is better on both benchmarks in Hindi, Tamil, and Telugu
(Fig.~\ref{fig:ablation}), reducing error by 10 to 39\% relative. The reduction appears on both IndicVoices
and the Vistaar sets, so it is not confined to spontaneous speech. Because elicitation is identical on
both sides, the difference cannot be attributed to topics, pairing, or
recording conditions. The gain
is largest in Telugu, at close to 40\% on both benchmarks, indicating that at
equal volume the gating, verification, and selection stages together account
for much of the accuracy of the resulting models.

\subsection{Do the gains transfer across architectures?}
\label{sec:sravaani}

To test whether the gains depend on Whisper, we continually fine-tune
SraVaani-1.0~\cite{sravaani2026}, which differs from Whisper in encoder
architecture, decoder formulation, and pretraining data, and is already
specialised to Indian languages. Continual fine-tuning on Monsoon reduces its
semantic WER on IndicVoices by 13 to 28\% relative across four representative
languages, Hindi, Bengali, Gujarati, and Telugu, with the largest gain in
Gujarati. On Hindi, where we evaluate the full suite, read-speech
benchmarks such as FLEURS and Kathbath show only marginal increases, of at
most 0.4 points, a regression that standard replay of the original training
data is designed to mitigate. The gains are therefore not specific to one
architecture, and they add accuracy on top of a model already trained on large
in-language data.

\subsection{How does Monsoon compare with commercial systems?}
\label{sec:commercial}

\begin{table}[t]
\centering
\caption{Semantic WER (\%) on IndicVoices. Best in bold. Commercial systems
evaluated in September 2026.}
\label{tab:commercial}
\small
\setlength{\tabcolsep}{3pt}
\begin{tabular}{@{}lccccccc@{}}
\toprule
System & hi & bn & gu & mr & ta & te & Avg. \\
\midrule
MAI-Transcribe-2     & \textbf{9.5} & \textbf{8.3} & \textbf{11.6} & \textbf{10.5} & \textbf{13.2} & 17.7 & \textbf{11.8} \\
Gemini 3.1 Pro       & 13.1 & 10.3 & 14.0 & 14.7 & 18.3 & 18.8 & 14.9 \\
ElevenLabs Scribe v2 & 13.2 & 12.0 & 16.6 & 16.4 & 22.7 & 23.2 & 17.4 \\
Muse Voice Transcribe & 16.2 & 14.5 & 31.9 & 24.8 & 35.2 & 33.0 & 25.9 \\
GPT Realtime Whisper & 33.4 & 45.0 & 47.9 & 48.1 & 64.5 & 55.0 & 49.0 \\
\midrule
Whisper medium       & 57.3 & 91.5 & 91.4 & 83.5 & 74.9 & 92.7 & 81.9 \\
\quad + Monsoon      & 11.8 & 10.0  & 12.3 & 12.2 & 15.6 & \textbf{16.1} & 13.0 \\
\bottomrule
\end{tabular}
\end{table}

Table~\ref{tab:commercial} places the fine-tuned model among commercial
systems on IndicVoices. Fine-tuning on Monsoon reduces Whisper medium's error
from 57--93\% to 10--16\%, after which it ranks second in five of the six languages, behind
only MAI-Transcribe-2, and first in Telugu, where it leads MAI-Transcribe-2 by 1.6 points, and it is ahead of Gemini 3.1 Pro,
ElevenLabs Scribe v2, Muse Voice Transcribe, and GPT Realtime Whisper in every
language. Averaged over the six languages, its error is 13\% lower than that of
Gemini 3.1 Pro, the second-best commercial system. The comparison is
conservative: the training data of the commercial systems is undisclosed and
may include the public IndicVoices training split, which the Monsoon model, a
769M-parameter model, has not seen.

\subsection{Does accuracy keep improving with more data?}
\label{sec:scaling}

\begin{figure}[t]
\centering
\includegraphics[width=\columnwidth]{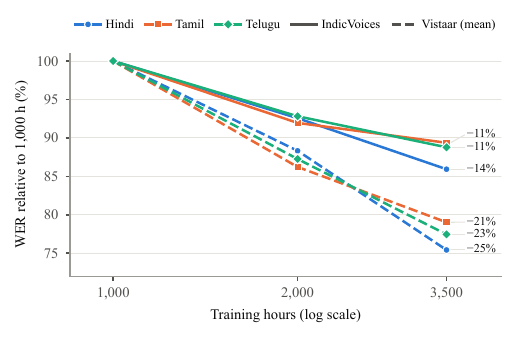}
\caption{Semantic WER relative to the 1{,}000-hour model, on IndicVoices
(solid) and the Vistaar mean (dashed). Labels: total relative reduction.}
\label{fig:scaling}
\end{figure}

We fine-tune Whisper medium on random subsets of 1{,}000 and 2{,}000 hours
and on the full corpus in Hindi, Tamil, and Telugu, holding the recipe fixed so that volume
is the only variable. Fig.~\ref{fig:scaling} expresses each
curve relative to its error at 1{,}000 hours. Error falls monotonically on all
six curves: from 1{,}000 hours to the full corpus, it drops by 11 to 14\%
relative on IndicVoices and by 21 to 25\% on the Vistaar sets, so the gain
from additional data is roughly twice as large on the Vistaar sets. The effect of volume
is consistent across the three languages. No curve has flattened at the full
collected volume, which suggests that collecting more data through the
framework would yield further improvements.

\section{Conclusion}
\label{sec:conclusion}

We presented Monsoon, a framework for collecting spontaneous speech for ASR
training at scale, and used it to build corpora of about 3{,}500 hours in each
of seven languages. At equal volume and with the same elicitation, its gating,
verification, and selection stages reduce error by 10 to 39\% relative over a
conventional protocol. Models trained on Monsoon match a read-speech baseline
on its own test sets and outperform it by 20 to 44 points on independently
collected spontaneous speech, the gains transfer to a second, Indic-specialised
architecture, and accuracy is still improving at the full collected volume.
A 10-hour sample of each of the 50 languages is available on request. Future work will extend the framework to further
languages.

\bibliographystyle{IEEEbib}
\FloatBarrier
\bibliography{strings,refs}

@misc{srivastav2025openasr,
  author       = {Srivastav, Vaibhav and others},
  title        = {Open {ASR} Leaderboard: Towards Reproducible and Transparent Multilingual and Long-Form Speech Recognition Evaluation},
  howpublished = {arXiv preprint arXiv:2510.06961},
  year         = {2025}
}

@inproceedings{radford2022whisper,
  author    = {Radford, Alec and Kim, Jong Wook and Xu, Tao and Brockman, Greg and McLeavey, Christine and Sutskever, Ilya},
  title     = {Robust Speech Recognition via Large-Scale Weak Supervision},
  booktitle = {Proc. ICML},
  year      = {2023}
}

@inproceedings{taguchi2024language,
  author    = {Taguchi, Chihiro and Chiang, David},
  title     = {Language Complexity and Speech Recognition Accuracy: Orthographic Complexity Hurts, Phonological Complexity Doesn't},
  booktitle = {Proc. ACL},
  pages     = {15493--15503},
  year      = {2024}
}

@inproceedings{ardila2020,
  author    = {Ardila, Rosana and others},
  title     = {{Common Voice}: A Massively-Multilingual Speech Corpus},
  booktitle = {Proc. LREC},
  pages     = {4218--4222},
  year      = {2020}
}

@inproceedings{wang2021,
  author    = {Wang, Changhan and others},
  title     = {{VoxPopuli}: A Large-Scale Multilingual Speech Corpus for Representation Learning, Semi-Supervised Learning and Interpretation},
  booktitle = {Proc. ACL-IJCNLP},
  pages     = {993--1003},
  year      = {2021}
}

@article{pratap2023,
  author  = {Pratap, Vineel and others},
  title   = {Scaling Speech Technology to 1,000+ Languages},
  journal = {Journal of Machine Learning Research},
  volume  = {25},
  number  = {97},
  pages   = {1--52},
  year    = {2024}
}

@inproceedings{bhogale2023,
  author    = {Bhogale, Kaushal Santosh and Sundaresan, Sai and Raman, Abhigyan and Javed, Tahir and Khapra, Mitesh M. and Kumar, Pratyush},
  title     = {{Vistaar}: Diverse Benchmarks and Training Sets for {Indian} Language {ASR}},
  booktitle = {Proc. Interspeech},
  pages     = {4384--4388},
  year      = {2023}
}

@misc{bhogale2026voiceindialargescalebenchmark,
  author       = {Bhogale, Kaushal and others},
  title        = {{Voice of India}: A Large-Scale Benchmark for Real-World Speech Recognition in {India}},
  howpublished = {arXiv preprint arXiv:2604.19151},
  year         = {2026}
}

@misc{javed2024indicvoices,
  author       = {Javed, Tahir and others},
  title        = {{IndicVoices}: Towards Building an Inclusive Multilingual Speech Dataset for {Indian} Languages},
  howpublished = {arXiv preprint arXiv:2403.01926},
  year         = {2024}
}

@misc{indicvoices_hf,
  author       = {{AI4Bharat}},
  title        = {{IndicVoices} Dataset Card},
  howpublished = {Hugging Face, updated Dec. 23, 2025. Accessed: Sep. 24, 2026. [Online]. Available: \url{https://huggingface.co/datasets/ai4bharat/IndicVoices}}
}

@misc{sravaani2026,
  author       = {Pulikodan, Sujith and others},
  title        = {{SraVaani 1.0}: Scaling Inclusive Speech Recognition for {Indic} Languages},
  howpublished = {arXiv preprint arXiv:2608.08235},
  year         = {2026}
}

@inproceedings{godfrey1992switchboard,
  author    = {Godfrey, John J. and Holliman, Edward C. and McDaniel, Jane},
  title     = {{SWITCHBOARD}: Telephone Speech Corpus for Research and Development},
  booktitle = {Proc. IEEE ICASSP},
  pages     = {517--520},
  year      = {1992}
}

@inproceedings{gales2014babel,
  author    = {Gales, Mark J. F. and Knill, Kate M. and Ragni, Anton and Rath, Shakti P.},
  title     = {Speech Recognition and Keyword Spotting for Low-Resource Languages: {Babel} Project Research at {CUED}},
  booktitle = {Proc. SLTU},
  pages     = {16--23},
  year      = {2014}
}

@inproceedings{javed2024lahaja,
  author    = {Javed, Tahir and others},
  title     = {{LAHAJA}: A Robust Multi-Accent Benchmark for Evaluating {Hindi} {ASR} Systems},
  booktitle = {Proc. Interspeech},
  year      = {2024}
}

@inproceedings{yong2025speakercount,
  author    = {Yong, Zheng-Xin and Pratap, Vineel and Auli, Michael and Maillard, Jean},
  title     = {Effects of Speaker Count, Duration, and Accent Diversity on Zero-Shot Accent Robustness in Low-Resource {ASR}},
  booktitle = {Proc. Interspeech},
  year      = {2025}
}

@inproceedings{reddy2021dnsmos,
  author    = {Reddy, Chandan K. A. and Gopal, Vishak and Cutler, Ross},
  title     = {{DNSMOS}: A Non-Intrusive Perceptual Objective Speech Quality Metric to Evaluate Noise Suppressors},
  booktitle = {Proc. IEEE ICASSP},
  pages     = {6493--6497},
  year      = {2021}
}

@inproceedings{babu2022xlsr,
  author    = {Babu, Arun and others},
  title     = {{XLS-R}: Self-Supervised Cross-Lingual Speech Representation Learning at Scale},
  booktitle = {Proc. Interspeech},
  pages     = {2278--2282},
  year      = {2022}
}

@article{kong2020panns,
  author  = {Kong, Qiuqiang and Cao, Yin and Iqbal, Turab and Wang, Yuxuan and Wang, Wenwu and Plumbley, Mark D.},
  title   = {{PANNs}: Large-Scale Pretrained Audio Neural Networks for Audio Pattern Recognition},
  journal = {IEEE/ACM Transactions on Audio, Speech, and Language Processing},
  volume  = {28},
  pages   = {2880--2894},
  year    = {2020}
}

@inproceedings{conneau2022,
  author    = {Conneau, Alexis and others},
  title     = {{FLEURS}: Few-Shot Learning Evaluation of Universal Representations of Speech},
  booktitle = {Proc. IEEE SLT},
  pages     = {798--805},
  year      = {2022}
}

@inproceedings{javed2023indicsuperb,
  author    = {Javed, Tahir and Bhogale, Kaushal Santosh and Raman, Abhigyan and Kumar, Pratyush and Kunchukuttan, Anoop and Khapra, Mitesh M.},
  title     = {{IndicSUPERB}: A Speech Processing Universal Performance Benchmark for {Indian} Languages},
  booktitle = {Proc. AAAI},
  pages     = {12942--12950},
  year      = {2023}
}

@inproceedings{diwan2021,
  author    = {Diwan, Anuj and others},
  title     = {{MUCS} 2021: Multilingual and Code-Switching {ASR} Challenges for Low Resource {Indian} Languages},
  booktitle = {Proc. Interspeech},
  pages     = {2446--2450},
  year      = {2021}
}

@misc{sarvam_llm_wer,
  author       = {{Sarvam AI}},
  title        = {{LLM-WER}: Semantic Evaluation for Automatic Speech Recognition},
  howpublished = {GitHub repository, 2026. Accessed: Sep. 1, 2026. [Online]. Available: \url{https://github.com/sarvamai/llm_wer}}
}

@misc{bhogale2026oiwer,
  author       = {Bhogale, Kaushal Santosh and others},
  title        = {Towards Orthographically-Informed Evaluation of Speech Recognition Systems for {Indian} Languages},
  howpublished = {arXiv preprint arXiv:2603.00941},
  year         = {2026}
}

\end{document}